\documentclass[conference]{IEEEtran}
\IEEEoverridecommandlockouts
\usepackage{cite}
\usepackage{amsmath,amssymb,amsfonts}
\usepackage{algorithm}
\usepackage{algpseudocode}
\usepackage{amsmath}
\usepackage{threeparttable}
\usepackage{booktabs}
\usepackage{multirow}
\usepackage{multicol}
\usepackage{pgfplots}
\usepackage{tabularx}
\usepackage{tikz}
\usepackage{graphicx}
\usepackage{textcomp}
\usepackage{xcolor}
\usepackage{hyperref}
\def\BibTeX{{\rm B\kern-.05em{\sc i\kern-.025em b}\kern-.08em
    T\kern-.1667em\lower.7ex\hbox{E}\kern-.125emX}}
\begin{document}

\title{A Machine Learning-Based Secure Face Verification
Scheme and Its Applications to Digital Surveillance
\thanks{This paper is accepted by International Conference on Digital Image and Signal Processing (DISP’19). The table of accepted papers can be found in \href{https://www.imm.dtu.dk/\string~alan/DISP2019_proceedings.pdf}{https://www.imm.dtu.dk/\string~alan/DISP2019\_proceedings.pdf}. This paper ID is 140.}}

\author{
\IEEEauthorblockN{1\textsuperscript{st} Huan-Chih Wang}
\IEEEauthorblockA{\textit{dept. Computer Science and Information Engineering} \\
\textit{National Taiwan University}\\
Taipei, Taiwan\\
whcjimmy@cmlab.csie.ntu.edu.tw}
\and
\IEEEauthorblockN{2\textsuperscript{nd} Ja-Ling Wu}
\IEEEauthorblockA{\textit{dept. Computer Science and Information Engineering} \\
\textit{National Taiwan University}\\
Taipei, Taiwan\\
wjl@cmlab.csie.ntu.edu.tw}
}

\maketitle

\begin{abstract}
Face verification is a well-known image analysis application and is widely used to recognize individuals in con- temporary society. However, most real-world recognition systems ignore the importance of protecting the identity-sensitive facial images that
are used for verification. To address this problem, we investigate how to implement a secure face verification system that protects the facial images from being imitated. In our work, we use the DeepID2 convolutional neural network to extract the features of a facial image and an EM algorithm to solve the facial verification problem. To maintain the privacy of facial images, we apply homomorphic encryption schemes to encrypt the facial data and compute the EM algorithm in the ciphertext domain. We develop three face verification systems for surveillance (or entrance) control of a local community based on three levels of privacy concerns. The associated timing performances are presented to demonstrate their feasibility for practical implementation.
\end{abstract}

\begin{IEEEkeywords}
Facial Verification, Machine Learning, Neural Networks, Encryption, Security, Digital Surveillance
\end{IEEEkeywords}

\section{Introduction}
\subsection{Problem Dfeinition}
In contemporary society, an individual’s identity sometimes needs to be verified before they are given permission to access some services or granted certain rights, for instance, a web user entering a password to login to websites or a person using his fingerprint to unlock his mobile phone. For modern applications, passwords, fingerprints, or even facial images are common ways to prove one’s identity. The main purpose of an identification check is to ensure a person is not disguised as someone else.

From an alternative perspective, the awareness of personal privacy protection has increased in recent years, and people have become more concerned about whether their personal information is being leaked to the public. Some biometric data, such as fingerprints and facial images, vary widely but are unique to each individual, so not only the data acquired after authorization but also the information used for authentication should be protected. In this way, personal data security can reach a new level.

Considering both identity authentication and personal data protection, in this work, we focused on building a secure facial authentication mechanism in which facial images can be protected and not be imitated by others. Clearly, identification becomes much more complicated if we protect the data from direct exploration. Analyzing the data in the encrypted domain and simultaneously maintaining high verification accuracy are the main challenges that we attempted to address.

\subsection{Contributions}
First, on the basis of DeepID2~\cite{sun2014deep}, we trained a convolutional neural network (CNN) model to extract facial image features and an EM model to verify whether two given images are taken from the same identity. Second, due to personal privacy concerns, we applied homomorphic encryption schemes to encrypt facial image features and move the face verification tasks to the ciphertext domain. Finally, we implemented an entrance guard system that allows residents living in a community to safely use their faces as the key to enter or exit.
\section{Background}
The three main techniques for face recognition are appearance-based, feature matching-based and hybrid approaches, as listed in~\cite{ramkumar2013face}. Appearance-based methods, such as eigenfaces and fisherfaces, take the whole face region as the raw input for recognition. Feature matching-based methods extract local templates, such as the nose, eyes and mouth, to represent a facial image and then use these templates for recognition. Two of the most famous approaches in this category are the hidden Markov model (HMM) and CNN. The third is a hybrid method that fuses the previous two methods together to improve recognition precision.

Currently, not every person is willing to share their sensitive data with others, so we need to consider privacy protection strategies, such as data encryption. Intuitively, if data are encrypted, they cannot be computed, searched or re-distributed. Fortunately, a new cryptography research area, called homomorphic encryption, enables direct computation on encrypted data. Decrypting the ciphertext domain computation as outcomes produces the same result as applying the computations in the plaintext domain. Therefore, homomorphic encryption approaches are applied to maintain data confidentiality and to enable useful applications directly on encrypted data.

Related works~\cite{sadeghi2009efficient,lagendijk2012encrypted,erkin2009privacy} focused on face identification in the encryption domain. They assumed that there is a database with a certain number of encrypted face images and that a person presents a query image and asks the system to determine whether the image is in the database. The Paillier cryptosystem was used to blind the query image so that the system user and/or manager could prevent information leakage during transmission and identification. For identification, the query image is decomposed into several eigenfaces, and the Euclidean distances between the query image and the face images stored in the database are computed. To decide whether the query image matches any images in the database, the server selects the best matching face, i.e., that which produces the smallest Euclidean distance.
\section{Preliminaries}
For the ease of explanation, we begin by reviewing some preliminary background knowledge.

\subsection{DeepID2 (Facial Feature Extraction)}
DeepID2 (Deep Identification-verification network version two)~\cite{sun2014deep} is a deep convolutional network that develops facial features from images. The structure of DeepID2 consists of four convolutional layers, and the first three convolutional layers are followed by a max-pooling layer. The input to DeepID2 is a 47 × 55 RGB image. The last layer is a 160-dimensional DeepID2 vector that is fully-connected to the third max-pooling layer, and it is in the last convolutional layer that the facial feature extraction is conducted. The main purpose of DeepID2 is not to classify face images but to extract useful facial features. DeepID2 uses two supervisory signals — an identification signal and a verification signal — to effectively represent facial features.

The identification signal is used to classify an input image into identity classes. Classification is achieved with an n-way softmax layer that outputs the probability distribution of each identity class. The softmax layer attempts to minimize the cross entropy loss, which represents the identification loss. The identification loss function of DeepID2 is defined as:
\begin{equation}
    Ident(f, t, \theta_{id})=\sum_{i=1}^{n}{-p_i \log{\hat{p_i}}}=-\log{\hat{p_t}},
\end{equation}
where $f$ denotes the DeepID2 vector, $t$ is the target class, and $\theta_{id}$ stands for the softmax later parameters. $p_i$ is the target probability distribution, where $p_i=0$ for all $i$, except $p_t=1$ for the target class $t$. $\hat{p_t}$ denotes the predicted probability distribution. 

Clearly, the identification signal is used to discriminate identity-related features from one another. The verification signal is used to gather the features that belong to the same identity closer together. This signal can effectively reduce intra-personal variation, which is commonly calculated based on the L1/L2 norm or the cosine similarity. The above loss function, with respect to the L2 norm, can be represented as:
\begin{multline}
    Verif(f_i, f_j, y_{ij}, \theta_{ve}) = \\ \left\{
    \begin{aligned}
        \frac{1}{2}||f_i-f_j||_{2}^{2}\text{ if }y_{ij}=1 \\
        \frac{1}{2}max(0, m-||f_i-f_j||_{2}^{2})\text{ if }y_{ij}=-1 \\
    \end{aligned}
    \right. .
\end{multline}

\subsection{EM Algorithm Model (Face Verification Algorithm)}
If we have a feature pair generated from two images, the challenge becomes how to check whether the two images correspond to the same identity. Baback Moghaddam et al. provided a successful face verification method in~\cite{moghaddam2000bayesian}. There are two hypotheses: $H_I$ states that the two images belong to the intra-personal class (i.e., the two images are generated from the same person) and $H_E$ states that the two images belong to the interpersonal class (i.e., the two images are generated by different people). Then, the maximum of a posterior rule can be written as a log-likelihood ratio, $r(x_1, x_2)$, between the conditional probabilities $P(\Delta|H_I)$ and $P(\Delta|H_E)$:
\begin{equation}
    r(x_1, x_2)=\log \frac{P(\Delta|H_I)}{P(\Delta|H_E)}.
\end{equation}

The above ratio can be seen as a measure of “how similar $x_1$ and $x_2$ are”. However, if the ratio is not sufficiently large or small, it is not easy to distinguish the intra-personal class from the interpersonal class according the value of $r(x_1, x_2)$. Therefore, other methods to estimate the likelihood ratio require further investigation.

\subsubsection{Reformulation of the Problem.} 
Dong Chen et al.~\cite{chen2012bayesian} proposed a reformulated face verification problem and used an EM algorithm to more precisely compute the likelihood ratio. First, we model the joint distribution of $x_1$ and $x_2$ as a Gaussian distribution; that is, $P(x_1, x_2|H_I)\sim N(0, \Sigma_I)$ and $P(x_1, x_2|H_E)\sim N(0, \Sigma_E)$, where $\Sigma_I$ and $\Sigma_E$ are the two corresponding covariance matrices that can be estimated from the intra-personal image pairs and the interpersonal image pairs, respectively. Then, a face image with identity $x$ can be represented as the sum of two independent Gaussian variables:
\begin{equation}
    x=\mu+\epsilon,
\end{equation}
where $\mu$ is the mean image of an identity and $\epsilon$ denotes the face variations (lighting, pose, and expression) corresponding to the same identity. If there are several different identities, each identity’s $\mu$ is different. If we focus on one specific identity’s images, each image $x_i$ can be represented as the mean image $\mu$ plus a slight variation image $\epsilon$. Here, we assume the latent variables have Gaussian distributions, that is, $\mu\sim N(0, S_\mu)$ and $\epsilon\sim N(0, S_\epsilon)$, where $\mathbf{S}_\mu$ and $\mathbf{S}_\epsilon$ are $d\times d$-dimensional matrices. Hereafter, we take above mentioned assumptions and representations as the prior information of a face.

\subsubsection{Joint Formulation with Prior Information.} 
Based on equation (4) and the assumption of statistical independence between $\mu$ and $\epsilon$, the covariance of two faces (or facial images) can be written as:
\begin{equation}
    cov(x_1, x_2)=cov(\mu_1, \mu_2)+cov(\epsilon_1, \epsilon_2).
\end{equation}
Under $H_I$, the identities of $\mu_1$ and $\mu_2$ are the same, but he variations $\epsilon_1$ and $\epsilon_2$ are independent. Based on equation (5), the covariance matrix of the distribution $P(x_1, x_2|H_I)$ can be written as:
\begin{equation*}
    \Sigma_I= 
    \begin{bmatrix}
        S_\mu+S_\epsilon & S_\mu \\
        S_\mu            & S_\mu+S_\epsilon.
    \end{bmatrix}
\end{equation*}

Under $H_E$, the identities of $\mu_1$ and $\mu_2$ and the variations $\epsilon_1$ and $\epsilon_2$ are both independent of each other. The covariance matrix of the distribution $P(x_1, x_2|H_E)$ can be written as:
\begin{equation*}
    \Sigma_E= 
    \begin{bmatrix}
        S_\mu+S_\epsilon & 0 \\
        0                & S_\mu+S_\epsilon.
    \end{bmatrix}
\end{equation*}

With the two covariance matrices $\Sigma_I$ and $\Sigma_E$, the log-likelihood ratio $r(x_1, x_2)$ can be written as:
\begin{equation}
\begin{split}
    r(x_1, x_2) &= \log\frac{P(x_1, x_2|H_I)}{P(x_1, x_2|H_E)} \\
                &= -[x_1,x_2]\Sigma_{I}^{-1}{[x_1,x_2]^T}+[x_1;x_2]\Sigma_{E}^{-1}{[x_1,x_2]^T} \\
                &= x_1^TAx_1+x_2^TAx_2-2x_1^TAx_2,
\end{split}
\end{equation}
where
\begin{equation*}
    A=(S_\mu+S_\epsilon)^{-1}-(F+G),
\end{equation*}
\begin{equation*}
    \Sigma_{I}^{-1}=
    \begin{pmatrix}
        S_\mu+S_\epsilon & S_\mu \\
        S_\mu            & S_\mu+S_\epsilon \\
    \end{pmatrix}^{-1} = 
    \begin{pmatrix}
        F+G & G \\
        G   & F+G \\
    \end{pmatrix},
\end{equation*}
and
\begin{align*}
    \Sigma_{E}^{-1} &=
    \begin{pmatrix}
        S_\mu+S_\epsilon & 0 \\
        0                & S_\mu+S_\epsilon \\
    \end{pmatrix}^{-1} \\ &= 
    \begin{pmatrix}
        (S_\mu+S_\epsilon)^{-1} & 0 \\
        G   & (S_\mu+S_\epsilon)^{-1} \\
    \end{pmatrix}.
\end{align*}

Both $\Sigma_{I}^{-1}$ and $\Sigma_{E}^{-1}$ are composed of $\mathbf{S}_\mu$ and $\mathbf{S}_\epsilon$. Therefore, we need
to compute $\mu$ and $\epsilon$ for all images in the training dataset to obtain matrices $\mathbf{S}_\mu$ and $\mathbf{S}_\epsilon$. In the next section, we use the EM algorithm and take $\mu$ and $\epsilon$ as latent variables to estimate the values of $\mathbf{S}_\mu$ and $\mathbf{S}_\epsilon$.

\subsubsection{EM Algorithm}
The EM algorithm~\cite{dempster1977maximum} is an iterative method used to estimate unobserved latent variables to determine the maximum likelihood or maximum a posteriori (MAP) ratio. The EM algorithm can be briefly divided into the E step and the M step. The E step creates an expectation function for the log-likelihood evaluation and calculates the likelihood value using the
current estimation of the unobserved latent parameters. The M step computes the latent parameters, which are used to maximize the expected log-likelihood found in the E step. The EM algorithm terminates when the latent parameters converge.

In our face verification problem, the $\mu$ and $\epsilon$ of each image are latent variables that we want to compute in the E step. At the
beginning of training, because we only have the value of each face $x_i$, we have to find a connection, which is an exception function of he image $x_i$ and the latent variables $\mathbf{S}_\mu$ and $\mathbf{S}_\epsilon$, between each images’ $\mu$ and $\epsilon$ in the E step. In the M step, we use the $\mu$ and $\epsilon$ of all the facial images to compute the covariance matrices $\mathbf{S}_\mu$ and $\mathbf{S}_\epsilon$ and then try to achieve convergence. The detailed execution of the EM algorithm is as follows.

\textbf{E Step:} Suppose that each identity is associated with $m$ images.
The relationships between the latent variables $h=[\mu,\epsilon_1,\cdots,\epsilon_m]$ and the face images $x=[x_1,\cdots,x_m]$ can be represented as:
\begin{equation*}
x=Ph,
\end{equation*}
where
\begin{equation*}
    P=
\begin{bmatrix}
    I & I & 0 & \cdots & 0 \\
    I & 0 & I & \cdots & 0 \\
    \vdots & \vdots & \vdots & \ddots & \vdots \\
    I & 0 & 0 & \cdots & I \\
\end{bmatrix}.
\end{equation*}

The distribution of $h$ is $h\sim N(0, \Sigma_h)$, where
\begin{equation*}
    \Sigma_h=
\begin{bmatrix}
    S_\mu & S_\mu & \cdots & S_\mu \\
    S_\epsilon & 0 & \cdots & 0 \\
    0 & S_\epsilon & \cdots & 0 \\
    0 & 0 & S_\epsilon & 0 \\
    0 & 0 & \cdots & S_\epsilon \\
\end{bmatrix}.
\end{equation*}

Because every image has the same mean $\mu$ but different variance $\epsilon$, we have $x\sim N(0, \Sigma_{x})$, where
\begin{equation*}
    \Sigma_x=
\begin{bmatrix}
    S_\mu+S_\epsilon & S_\mu & \cdots & S_\mu \\
    S_\mu & S_\mu+S_\epsilon & \cdots & S_\mu \\
    \vdots & \vdots& \ddots& \vdots \\
    S_\mu & S_\mu & \cdots& S_\mu+S_\epsilon \\
\end{bmatrix}.
\end{equation*}

Given the observation $x$, the expection of the hidden variable $h$ is:
\begin{equation}
\begin{split}
    E(h|x) &= P^{-1}x \\
           &= P^{-1}\Sigma_x\Sigma_x^{-1}x \\
           &= P^{-1}P\Sigma_hP^T\Sigma_x^{-1}x \\
           &= \Sigma_hP^T\Sigma_x^{-1}x.
\end{split}
\end{equation}

Assume $\Sigma_x$ is invertible, we let
\begin{equation*}
    \Sigma_x^{-1}=
\begin{bmatrix}
    F+G & G & \cdots & G \\
    G   & F+G & \cdots & G \\
    \vdots & \vdots & \ddots & \vdots \\
    G & G & \cdots & F+G \\
\end{bmatrix}.
\end{equation*}

Since $\Sigma_x\Sigma_x^{-1}=I$, we can obtain the following two equations:
\begin{equation*}
    \left\{
    \begin{aligned}
        (S_\mu+S_\epsilon)(F+G)+(m-1)S_\epsilon G= I\\
        (S_\mu+S_\epsilon)G + S_\mu F +(m-1)S_\epsilon G= 0\\
    \end{aligned}
    \right. .
\end{equation*}

Soving the equations above, we get $F=S_\epsilon^{-1}$ and $G=-(mS_\mu+S_\epsilon)^{-1}S_\mu S_\epsilon^{-1}$.

As a result, for each identity, the mean is:
\begin{equation}
    \mu=\sum_{i=1}^{m}{(F+mG)x_i},
\end{equation}
and the variance associated with face $x_j$ is:
\begin{equation}
    \epsilon=x_j+\sum_{i=1}^{m}{S_\epsilon Gx_i}.
\end{equation}

Based on equations (8) and (9) and the above discussions, we can calculate the mean $\mu$ for all identities and $\epsilon$ for each face image with respect to the mean of its identity.

\textbf{M Step:} In the M step, we update $S_\mu$ and $S_\epsilon$ using the latent variables $\mu$ and $\epsilon$, where 
\begin{equation*}
S_\mu=\frac{1}{n}\sum_{i}{\mu_i\mu_i^T}
\end{equation*}
and 
\begin{equation*}
S_\epsilon=\frac{1}{n}\sum_{i}{\epsilon_{ij}\epsilon_{ij}^T}.
\end{equation*}

The EM algorithm stops when $S_\mu$ and $S_\epsilon$ are both converge.

\subsection{Encryption Mechanisms}
\subsubsection{Homomorphic Encryption}
Homomorphic encryption is a class of encryption algorithms in which certain operations can be directly carried out on ciphertexts to generate the corresponding encrypted results. After decrypting the encrypted result, it will match the result of the same operations performed on the corresponding plaintexts. We use $m$ to denote a plaintext message and $[m]$ to denote the corresponding ciphertext. As a result, $[m]=Enc_{PK}(m)$, which means $m$ is encrypted with public key PK, and $m=Dec_{PK}([m])$, which means $[m]$ can be decrypted with the secret key SK.

If homomorphic encryption scheme works well with both addition and multiplication, it is called a fully homomorphic encryption (FHE) scheme, while other method are called somewhat fully homomorphic encryption (SFHE) schemes. In this work, the high performance and easy-to-implement Fan and Vercauteren~\cite{fan2012somewhat} FHE scheme (FV-scheme) is adopted.

\subsubsection{Paillier Cryptosystem}
The Paillier cryptosystem~\cite{paillier1999public} is an SFHE algorithm that has an additional operation feature and a multiplication operation with limit. The whole cryptosystem is based on the face that it is difficult to compute the n-th residue class. In addition to being an additive homomorphism, the Paillier scheme can be computed rapidly; therefore, it is wildly used in the fields of data security and privacy preservation.

\subsection{Secure Protocols}
In this section, two secure computational protocols involved in our proposed system are described, briefly.

\subsubsection{Secure Comparison Protocol}
Sometimes, we have two encrypted numbers to compare. In this work, we slightly alter Veugen’s protocol~\cite{veugen2011comparing} to achieve our goal. Assume Bob has two encrypted numbers $[a]$ and $[b]$, both with bit length $l$ in the plaintext domain, which are encrypted with Alice’s public key. Assume Alice wants to know whether $b$ is larger than $a$. Because Alice has the secret key, Bob cannot directly send the data to Alice and ask for the comparison result. As a consequence, Alice and Bob have to work together. From beginning to end, neither Alice nor Bob knows the actual values of $a$ and $b$. The main idea is to determine the sign of the most significant bit (MSB) of $x=b+2^l-a$, whose polarity indicates whether $a\leq b$.
\begin{algorithm}
    \caption{The Secure Comparison Protocol} \label{alg:comparison}
    \begin{algorithmic}[1]
    \Require Alice ($\mathcal{A}$) owns PK and SK. Bob ($\mathcal{B}$) owns $[a]$, $[b]$, and PK.
    \Ensure Alice gets bit $t$, where $t=1$ if $a \leq b$.
    \State $\mathcal{B}$ computes $[x]=[b]+[2^l]-[a]$ and $[z]=[x]+[r]$.
    \State $\mathcal{B}$ sends $[z]$ to $\mathcal{A}$.
    \State $\mathcal{A}$ decrypts $[z]$.
    \State $\mathcal{A}$ computes $d=z\ mod\ 2^l$, and $\mathcal{B}$ computes $c=r\ mod\ 2^l$.
    \State $\mathcal{A}$ and $\mathcal{B}$ privately use DGK protocol~\cite{damgard2009correction} to let $\mathcal{B}$ gets the encrypted bit $[t']$ such that $t'=(d < c)$, which means $t'=1$ iff $d<c$ is true.
    \State $\mathcal{A}$ encrypts $z_{l+1}$ to be $[z_{l+1}]$, and $\mathcal{B}$ encrypts $r_{l+1}$ to be $[r_{l+1}]$.
    \State $\mathcal{A}$ sends $[z_{l+1}]$ to $\mathcal{B}$.
    \State $\mathcal{B}$ computes $[t]=[t']\dot[z_{l+1}]\dot[r_{l+1}]$.
    \State $\mathcal{B}$ sends $[t]$ to $\mathcal{A}$.
    \State $\mathcal{A}$ decrypts $[t]$ and gets the plaintext $t$.
    \end{algorithmic}
\end{algorithm}

\subsubsection{Secure Matrix Multiplication Protocol}
In our system, we require a secure computation protocol to handle the matrix multiplications. This protocol is improved from the “Conducting dot products” protocols presented in~\cite{bost2014machine}. Suppose Alice has an $m\times n$ matrix and Bob has an $n\times p$ matrix. Alice encrypts her matrix, sends the result to Bob and asks Bob to compute the multiplication result of the two matrices.
\begin{algorithm}
    \caption{The Secure Matrix Multiplication Protocol} \label{alg:comparison}
    \begin{algorithmic}[1]
    \Require Alice ($\mathcal{A}$) has a matrix $X\in \mathbb{Z}^{m\times n}$. Bob has a matrix $Y\in \mathbb{Z}^{n\times p}$.
    \Ensure Bob gets the encrypted matrix $[XY]$.
    \State $\mathcal{A}$ encrypts the matrix $X$ with PK to obtain $[X]$.
    \State $\mathcal{A}$ sends $[X]$ to $\mathcal{B}$.
    \State $\mathcal{B}$ computes $[XY]_{ij}=\prod_{r}{[X_{ir}]^{Y_{rj}}}$ using Paillier or $[XY]_{ij}=\sum_{r}{[X_{ir}][y_{rj}]}$ if an HE scheme (such as the FV scheme) is involved.
    \end{algorithmic}
\end{algorithm}

\subsubsection{Dataset}
In this work, two datasets are used to perform and/or testing. The first is the “Labeled Faces in the Wild (LFW)” dataset, which was published in~\cite{huang2008labeled}. LFW contains more than 13,000 images from the web in which approximately 1700 subjects have more than two images. This dataset is used as the benchmark for our classification experiments. The second dataset is CASIA-WebFace~\cite{yi2014learning}. An organization called CBSR (Center for Biometrics and Security Research) in China collected the images, which contain 10,575 subjects in 494,414 images taken from the Internet. The CASIA dataset is now the largest face-related public dataset. CASIA is used to train DeepID2 and compute matrices A and G, which are needed to build the EM model.
\section{The Proposed Schemes}
\subsection{Overview}
As previously described, we first use the CASIA-WebFace dataset to train DeepID2 and then apply it to establish the EM model in the plaintext domain. As shown in Figure 4.1, DeepID2 is used to extract the required features from two selected image. Each input to DeepID2 is an image cropped with the appropriate preprocessing tool, and the network outputs a 160-dimensional image feature vector. Therefore, we run DeepID2 two times. Then, the EM model is applied to compute the likelihood ratio of the two feature vectors obtained in the above step. In the last step, we compare the likelihood ratio with a given threshold to determine whether the two images represent the same person.
\begin{figure}[htb]
    \centering
    \includegraphics[width=1\columnwidth]{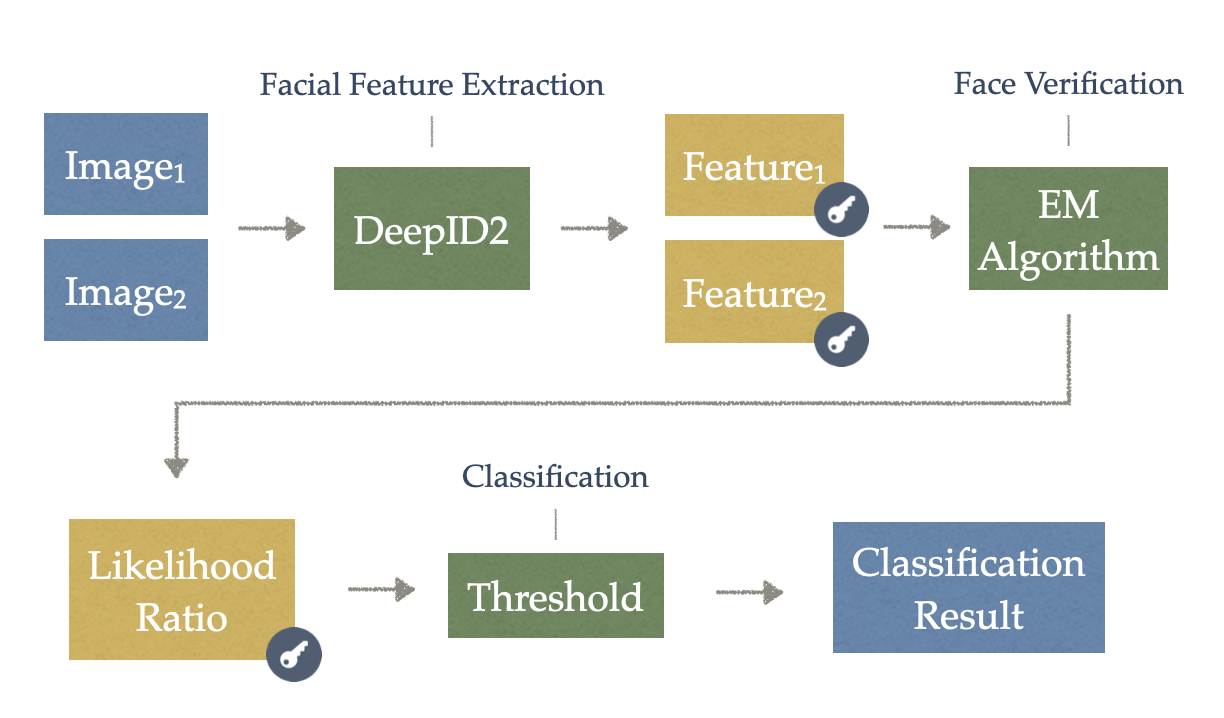}
    \caption{Overall workflow of the facial verification process.} 
    \label{verification_workflow}
\end{figure}

In the above discussions, all the steps from the input images to the classification result are conducted in the plaintext domain. However, if the set of images is highly sensitive, it should be protected to satisfy certain security concerns. That is, the features extracted by DeepID2 should be encrypted, and from then on, the likelihood ratio and even the classification result also have to be encrypted. As a result, as shown in Fig~\ref{verification_workflow}, the computation associated with the EM modeling and the comparison with a given threshold need to be redesigned in consideration of privacy concerns. In our work, we perform the secure EM modeling computation by using the secure matrix multiplication protocol and conduct the secure comparison task by using the secure comparison protocol.

Note that, depending on the application scenario, the two feature vectors may not both need to be encrypted. We provide three application scenarios based on different degrees of privacy concerns that will be discussed in the next section.

\subsection{Application Scenarios}
In this section, we use the above mentioned components to develop three face recognition systems for surveillance or entrance control of a local community under different scenarios.

In all scenarios, we aim to build a security system using residents’ faces as the key to access their entrance gate. All application scenarios have two operational stages. The first is the registration stage, which occurs only when a new householder moves into the community. The newcomer has to take a photo that becomes one of the images used for facial verification. This photo is stored on a server. Recall that the face verification module judges whether two images are of the same individual. Therefore, the registered images are used whenever a householder comes to the community’s entrance gate. After a resident completes registration, he is given an ID card that records an encrypted ID number, which varies from person to person. This number is then presented when the resident returns to the entrance gate.

The second stage is called the verification stage, which occurs every time a resident wants to enter the entrance gate. In this stage, a photo of the resident is immediately captured at the gate and sent to the server with his ID number. The server uses the ID number to select the corresponding photo from the registration stage. After the photo corresponding to the ID number is received, the next step is to compute the likelihood ratio between the two photos. Finally, the client and the server run a secure comparison protocol, where the client is Alice and the server is Bob, to determine whether the resident is disguised. If the identification system finds that the two images come from the same individual, the community’s security system lets the person in; otherwise, the person is locked outside the gate.

In real applications, as shown in Figure 4.2, the client-site is equipped with a set of peripheral devices (e.g., webcam, microcomputer and card reader) to compute mathematical operations, i.e., it is able to run image preprocessing and CNN models, as well as encryption and decryption. Webcams need to be installed in two locations. The first is in the community’s security room because each new resident needs to register as a
new member of this community. After registration, a microcomputer writes the ID number into a new ID card. The second location is along the front of the entrance gate or at the front door of each house. When a resident wants to enter the entrance gate or his home, he first plugs his ID into the card reader and then conducts the verification stage, with his face, by himself. If the verification is successful, the microcomputer can unlock the door and let the resident in. Note that in our second and third scenarios, since residents want to protect their data from the server, they receive a key-pair from a trusted third party, the authority of the community’s security center or a designated security service firm.
\begin{figure}[htb]
    \centering
    \includegraphics[width=1\columnwidth]{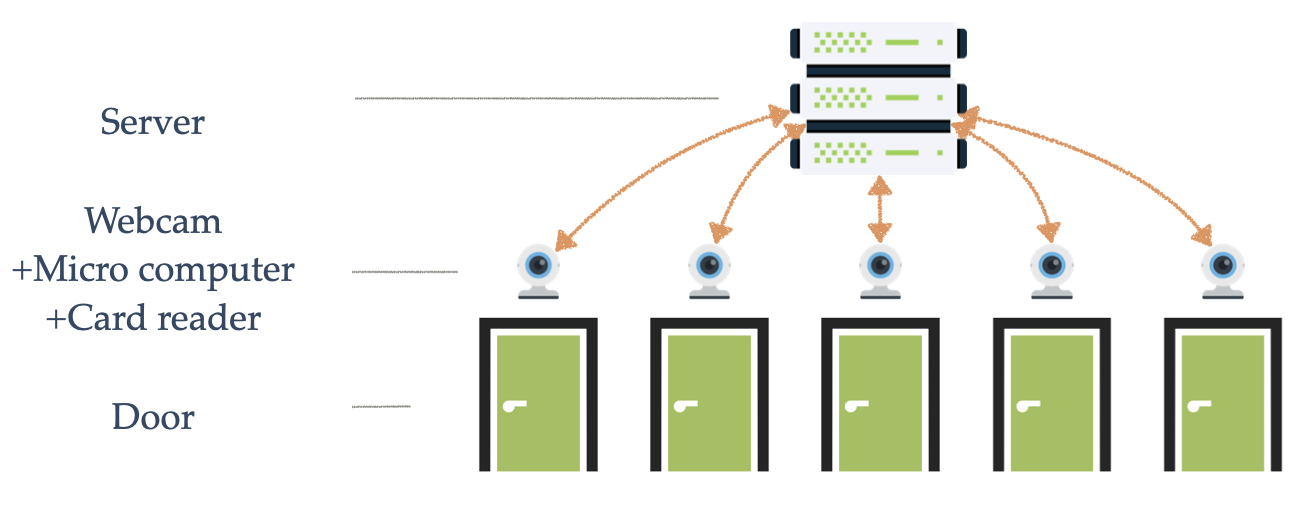}
    \caption{The structure of a community security system.} 
    \label{verification_workflow}
\end{figure}

\subsubsection{The First Scenario}
In this scenario, the server is a trusted security service firm, so clients, i.e., residents, can store their plaintext facial features in the company’s database. Because protecting the safety of transmission is the focus, in both the registration and verification stages, a resident receives a public key issued by the security company that can be used to encrypt the feature vectors to prevent their face-related information from being stolen by hackers during transmission. After the security company receives the encrypted features, the server decrypts those features for further processing, whether by storing features in a database or computing the required likelihood ratio. Because the server is responsible for decrypting the features, all computations in the verification stage are performed in the plaintext domain; therefore, this scenario is completed the fastest among all three scenarios. In the last step of the verification stage, the server first encrypts the likelihood ratio and the threshold and then applies the secure comparison protocol with the aid of the client to obtain the comparison result in the plaintext domain. Algorithm 3 and Algorithm 4 are used to perform the registration and verification stages in the first scenario, respectively.
\begin{algorithm}
    \caption{The Registration State of the Scenario 1} \label{alg:comparison}
    \begin{algorithmic}[1]
    \Require The webcam ($\mathcal{W}$) does nothing, while the server ($\mathcal{S}$) generates matrices $A$ and $G$ for the EM modeling and the public key PK$_{\mathcal{S}}$.
    \Ensure The new resident ($\mathcal{C}$) gets an ID card that records the encrypted index $[id]_\mathcal{S}$. The Server computes $x^TAx$ and $Gx$.
    \State $\mathcal{W}$ takes a photo for the new resident and extracts the feature $x$.
    \State $\mathcal{W}$ encrypts the vector $x$.
    \State $\mathcal{W}$ sends $[x]_{\mathcal{S}}$ to $\mathcal{S}$.
    \State $\mathcal{S}$ decrypts $[x]_{\mathcal{S}}$.
    \State $\mathcal{S}$ computes $x^TAx$ and $Gx$ and stores them in a database.
    \State $\mathcal{S}$ produces an ID card associated with an encrypted ID number $[id]_{\mathcal{S}}$.
    \State $\mathcal{S}$ sends the ID card to $\mathcal{C}$.
    \State $\mathcal{C}$ receives the ID card.
    \end{algorithmic}
\end{algorithm}
\begin{algorithm}
    \caption{The Verification State of the Scenario 1} \label{alg:comparison}
    \begin{algorithmic}[1]
    \Require The webcam's card reader ($\mathcal{WR}$) reads the ID card. The server generates matrices $A$ and $G$, a given threshold for checking the EM model, PK$_{\mathcal{S}}$, SK$_{\mathcal{S}}$, and the database storing $x_i^TAx_i$ and $Gx_i$.
    \Ensure The webcam gets the plaintext classification result.
    \State $\mathcal{W}$ takes a photo for the upcoming resident and extracts the feature $y$.
    \State $\mathcal{W}$ encrypts $y$ and $\mathcal{WR}$ reads $[id]_{\mathcal{S}}$.
    \State $\mathcal{W}$ sends $[y]_{\mathcal{S}}$ and $[id]_{\mathcal{S}}$ to $\mathcal{S}$.
    \State $\mathcal{S}$ decrypts $[y]_{\mathcal{S}}$ and $[id]_{\mathcal{S}}$.
    \State $\mathcal{S}$ uses $id$ to find the corresponding identity's $x^TAx$ and $Gx$ from the database.
    \State $\mathcal{S}$ computes $LR=x^TAx+y^TAy-2y^TGx$.
    \State $\mathcal{W}$ and $\mathcal{S}$ run a secure comparison protocol and $\mathcal{W}$ gets the classification result and controls whether the gate opens.
    \end{algorithmic}
\end{algorithm}

\subsubsection{The Second Scenario}
In the second scenario, we not only protect the transmission security (as in the first scenario) but also ensure the safety of images stored in the database. In this scenario, the server is a local security monitoring system that the security service firm placed in the security guard room of the community. Because the system is not located inside the security company, the residents do not trust the data security claimed by the company. From a resident’s point of view, the data might be stolen by thieves (both insiders and outsiders), so the database should be encrypted. Therefore, the photos taken during the registration stage should be encrypted due to the above-mentioned privacy and data security concerns.

Let us take a close look at the computation of $x^TAx$ in the likelihood ratio equation (6); it will firstly multiply $x$ by $A$ and then $x^TA$ by $x$. If $x$ is an encrypted vector, every element in $x^TA$ will also be in the ciphertext domain. Furthermore, because $x^TA$ and $x$ are now both in the ciphertext domain, the computation of $x^TAx$ cannot be performed directly using the Paillier cryptosystem. To reduce the execution time, the Kronecker product is used to let the likelihood computation remain computable under the Paillier cryptosystem. That is,
$vec(x^TAx)=(x^T\otimes x)vec(A)$, where $\otimes$ denotes the vector
Kronecker product and “T” denotes the transpose of vectors.

Therefore, the client-site is asked to compute $x^T\otimes x$ in advance
and encrypts both $x^T\otimes x$ and $x$ and then sends $[x^T\otimes x]$ and $[x]$ to the server for storage and further computation. Because $x$ is encrypted, to compute the likelihood ratio, the server relies on the previously described secure matrix multiplication protocol. Furthermore, to compare the encrypted likelihood ratio with the given threshold, the server requires help from the above-mentioned secure comparison protocol, where the server plays the role of Bob and the client plays the role of Alice. Let $[x]_C$ denote that vector $x$ is encrypted with the client’s public key and $[x]_S$ denote $x$ is encrypted with the server’s public key.

Algorithm 5 and Algorithm 6 are used to perform the registration
and verification stages in the second scenario.
\begin{algorithm}
    \caption{The Registration State of the Scenario 2} \label{alg:comparison}
    \begin{algorithmic}[1]
    \Require The webcam ($\mathcal{W}$) does nothing, while the server ($\mathcal{S}$) generates matrices $A$ and $G$ for the EM modeling and the public key PK$_{\mathcal{S}}$ and PK$_{\mathcal{C}}$.
    \Ensure The new resident ($\mathcal{C}$) gets an ID card recording the encrypted index $[id]_\mathcal{S}$. The server computes $[x^T\otimes x]_\mathcal{C}A$ and $G[x]_\mathcal{C}$.
    \State The same as step 1 in Algorithm 3.
    \State $\mathcal{W}$ computes $x^T\otimes x$ and encrypts $x^T\otimes x$ and $x$.
    \State $\mathcal{W}$ sends $[x^T\otimes x]_\mathcal{C}$ and $[x]_\mathcal{C}$ to $\mathcal{S}$.
    \State $\mathcal{S}$ computes $[x^TAx]_\mathcal{C}A$ and $G[x]_\mathcal{C}$ and stores them in a database.
    \State The same as steps 6-8 in Algorithm 3.
    \end{algorithmic}
\end{algorithm}
\begin{algorithm}
    \caption{The Verification State of the Scenario 2} \label{alg:comparison}
    \begin{algorithmic}[1]
    \Require The webcam's card reader ($\mathcal{WR}$) reads the ID card of a resident to get $[id]_\mathcal{S}$. The server generates matrices $A$ and $G$, a given threshold for checking the EM model, PK$_{\mathcal{S}}$, PK$_{\mathcal{C}}$, and SK$_{\mathcal{S}}$, and the database storing $[x_i^TAx_i]_{\mathcal{C}_i}$ and $[Gx_i]_{\mathcal{C}_i}$.
    \Ensure The webcam gets the plaintext classification result.
    \State $\mathcal{W}$ takes a photo for the upcoming resident and extracts the feature $y$.
    \State $\mathcal{WR}$ reads $[id]_{\mathcal{S}}$.
    \State $\mathcal{W}$ computes $y^T\otimes y$ and encrypts $y^T\otimes y$ and $y$.
    \State $\mathcal{W}$ sends $[y^T\otimes y]_\mathcal{C}$, $[y]_\mathcal{C}$ and $[id]_\mathcal{S}$ to $\mathcal{S}$.
    \State $\mathcal{S}$ decrypts $[id]_{\mathcal{S}}$ and uses $id$ to find the corresponding identity's $[x^TAx]_\mathcal{C}A$ and $[Gx]_\mathcal{C}$ from the database.
    \State $\mathcal{S}$ computes $y^TAy$ and encrypts it with PK$_{\mathcal{C}}$.
    \State $\mathcal{S}$ computes $LR=[x^TAx]_\mathcal{C}A+[y^TAy]_\mathcal{C}-2y^T[Gx]_\mathcal{C}$.
    \State The same as step 7 in Algorithm 4.
    \end{algorithmic}
\end{algorithm}

\subsubsection{The Third Scenario}
Since an FHE scheme usually costs too much time to compute, it is not a good choice in practice. However, with the aid of FHE cryptosystems, we can design the most secure protocol, which is far safer than the above protocols. In this scenario, the server is also an untrusted security system, and we try to protect all the residents’ privacy, including image features stored on the database and the image features sent to the system for verification.

Algorithm 7 and Algorithm 8 are used to perform the registration and verification stages in the third scenario.
\begin{algorithm}
    \caption{The Registration State of the Scenario 3} \label{alg:comparison}
    \begin{algorithmic}[1]
    \Require The same as that of Algorithm 5.
    \Ensure The new resident ($\mathcal{C}$) gets an ID card recording the encrypted index $[id]_\mathcal{S}$. The server computes $[x]_\mathcal{C}^TA[x]_\mathcal{C}$ and $G[x]_\mathcal{C}$.
    \State The same as that of  Algorithm 5.
    \State $\mathcal{W}$ encrypts $x$.
    \State $\mathcal{W}$ sends $[x]_\mathcal{C}$ to $\mathcal{S}$.
    \State $\mathcal{S}$ computes $[x]_\mathcal{C}^TA[x]_\mathcal{C}$ and $G[x]_\mathcal{C}$ and stores them in a database.
    \State The same as steps 6-8 in Algorithm 3.
    \end{algorithmic}
\end{algorithm}
\begin{algorithm}
    \caption{The Verification State of the Scenario 3} \label{alg:comparison}
    \begin{algorithmic}[1]
    \Require The webcam's card reader ($\mathcal{WR}$) reads the ID card of a resident to get $[id]_\mathcal{S}$. The server generates matrices $A$ and $G$, a given threshold for checking the EM model, PK$_{\mathcal{S}}$, PK$_{\mathcal{C}}$, and SK$_{\mathcal{S}}$, and the database storing $[x]_\mathcal{C}^TA[x]_\mathcal{C}$ and $G[x]_\mathcal{C}$.
    \Ensure The webcam gets the plaintext classification result.
    \State The same of step 1 to step 4 in Algorithm 4.
    \State $\mathcal{S}$ uses $id$ to find the corresponding identity's $[x]_\mathcal{C}^TA[x]_\mathcal{C}$ and $G[x]_\mathcal{C}$ from the database.
    \State $\mathcal{S}$ computes $LR=[x]_\mathcal{C}^TA[x]_\mathcal{C}+[y]_\mathcal{C}^TA[y]_\mathcal{C}-2[y]_\mathcal{C}^TG[x]_\mathcal{C}$.
    \State The same as step 7 in Algorithm 4.
    \end{algorithmic}
\end{algorithm}

\subsection{A General Comparison of the Three Scenarios}
We perform a simple comparison of the three scenarios in Table~\ref{tab_comparisons}. The comparison results can be used to select the most appropriate algorithms.
\begin{table}[htb]
\centering
\caption{Comparisons among the Three Scenarios.}\label{tab_comparisons}
\begin{tabularx}{\columnwidth}{lr>{\raggedleft\arraybackslash}X}
    \toprule
    Scenario & Encryption Type & Characteristics \\
    \midrule
    1 & Plaintext & Fastest \\
    2 & SFHE (Paillier) & Only protects data stored on database \\
    3 & FHE (FV Scheme) & Protects all images, but it is slow. \\
    \bottomrule
\end{tabularx}
\end{table}

\section{Performance Evalaution}
Timing and accuracy are the two main factors we consider for a useful verification system. Some experiments are conducted to examine the corresponding costs of the secure face recognition schemes. The following experiments are all conducted on a personal computer with a Linux Mint 17.2 operating system and an i7-2600 CPU with 4 GB RAM.
\subsection{Accuracy}
In this experiment, we compare (in terms of verification accuracy) DeepID2 with the HD-LBP (high-dimensional local binary patterns) scheme proposed in~\cite{chen2013blessing}. HD-LBP is a widely adopted facial feature extraction method that combines multi- scale patches and an LBP method to extract image features. The HD-LBP scheme extracts a 2000-dimensional feature vector, and PCA (principle component analysis)~\cite{dunteman1989principal} is used to reduce the feature dimensions to 160, identical to DeepID2. We also report the face verification accuracy results with respect to the LFW dataset. We randomly choose 3000 image pairs from different individuals and 3000 image pairs from the same individuals. As shown in Figure 5.1, the maximum verification accuracy of the HD-LBP is 89\%, which is approximately 7\% less than that of DeepID2.
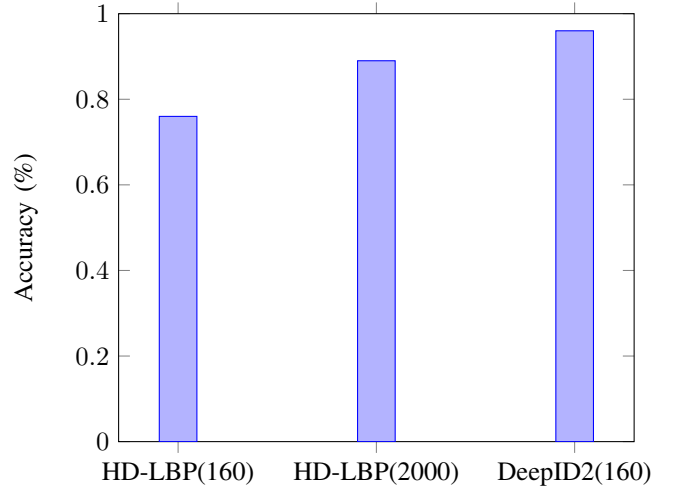
\begin{figure}
    \centering
    \begin{tikzpicture}
        \begin{axis}[
            ybar,
            ylabel={Accuracy (\%)},
            xtick={1, 2, 3},
            xticklabels={HD-LBP(160), HD-LBP(2000), DeepID2(160)},
            ymin=0, ymax=1,
            bar width=0.5cm,
            enlarge x limits=0.15,
        ]
        \addplot coordinates {(1,0.76) (2,0.89) (3,0.96)};
        \end{axis}
    \end{tikzpicture}
    \caption{Accuracy comparison between the HD-LBP and DeepID2 based schemes.}
\end{figure}

\subsection{Timing Performance}
In the proposed scenarios, we implement our face verification system based on the open-sourced Paillier code written by Raphael Bost~\cite{ciphermed} and FV-NFLlib written by CryptoExperts~\cite{fvnfllib}. Both have been accelerated by OpenMP. We examine the execution time required by different components of the protocols with different types of cryptosystems. Furthermore, we estimate the required execution times in different stages. For the Paillier cryptosystem, in the registration stage, we examine the execution time for encrypting $x^T\otimes x$ and $x$ vectors at the client side and that of $[x^T\otimes x]_C$ and $G[x]_C$ at the server side. In verification stage, we examine the execution tie for encryption vector $y$ at the client side and that of the likelihood ratio, which includes the computation of $y^TAy$ and $y^TG[x]_C$, summing up $[x^T\otimes x]A$, $[y^TAy]_C$ and $-2y^TG[x]_C$, at the server side.

In experiments with FHE, the client in both the registration stage and the verification stage needs to encrypt a 160-dimensional vector. In the registration stage, the server computes $[x]_C^{T}A[x]_C$ and $G[x]_C$, and in the verification stage, the server needs not only to compute $[y]_C^T A[x]_C$ and $G[y]_C$ but also to sum $[x]_C^{T}A[x]_C$, $[y]_C^T A[x]_C$ and $-2[x]_C G[y]_C$.

As shown in Table 5.1, the approach without any encryption scheme is the fastest. Although the application with the Paillier cryptosystem costs substantial time in the registration stage, it is acceptable because the registration stage occurs only once for each new resident. In addition, the verification stage with the Paillier cryptosystem can be completed in nearly real time, so a resident does not need to wait in front of the gate for a long time. Lastly, the server needs to run for more than two minutes in the registration stage and more than one minute in the verification stage when an FHE scheme is used. Although the server runs for a long time, it is necessary if data security and/or user privacy is a requirement.

\begin{table}[htb]
\centering
\caption{Execution Times (in seconds) of the proposed schemes.}\label{tab_comparisons}
\begin{tabular}{lrrrrr}
    \toprule
    Stage & Party & Plaintext & Paillier & FHE \\
    \midrule
    \multirow{2}{*}{Registration Stage} & Client & 0.003 & 4.409 & 0.86 \\
    \cmidrule{2-5}
    & Server & 0.014 & 13.233 & 160 \\
    \midrule
    \multirow{2}{*}{Verification Stage} & Client & 0.003 & 0.031 & 0.86 \\
    \cmidrule{2-5}
    & Server & 0.016 & 0.052 & 79 \\
    \bottomrule
\end{tabular}
\end{table}
\section{conclusion}
By implementing the good feature extraction performance of DeepID2, we develop a face verification system that reaches 96\% accuracy. Due to advances in homomorphic encryption protocols, we can directly transfer a face verification system from the plaintext domain into the ciphertext domain. Furthermore, we analyze three different types of scenarios based on different privacy concerns to conduct surveillance and/or entrance control for a community. Although we use only one patch to extract face features, in the near future, we will attempt to use multi-patches to obtain more precise features. In addition, we will try to accelerate the execution of the encryption domain computations, specifically for FHE schemes.

\section{Acknowledge}
This research is supported by Minister of Science and Technology, Taiwan ROC, under the contract number : MOST 106-3114-E-002-009. This article is rewritten based on my master thesis.

\bibliographystyle{IEEEtran}
\bibliography{main}

\end{document}